\documentclass[conference]{IEEEtran}
\IEEEoverridecommandlockouts    

\usepackage{multirow}
\usepackage{lipsum}
\usepackage{amsmath}
\usepackage{amssymb}
\usepackage[ruled,vlined]{algorithm2e}
\usepackage{graphicx}
\graphicspath{{./Figures/}}

\title{\vspace{5mm}
\LARGE \bf Integrating Traffic Noise Emission Modelling \\ into Variable Speed Limit Control}

\author{
Jiawen Meng$^{1}$,
John Pravin Arockiasamy$^{1}$,
Alexey Vinel$^{1,2}$\\[1mm]
\small
$^{1}$Institute AIFB, Karlsruhe Institute of Technology, Germany\\
$^{2}$School of Information Technology, Halmstad University, Sweden\\
\thanks{This paper has been accepted for presentation at the IEEE Intelligent Transportation Systems Conference (ITSC) 2026.}
}
\begin{document}
\maketitle
\thispagestyle{empty}
\pagestyle{empty}

\begin{abstract}

Road traffic noise remains a major environmental challenge, yet most speed management strategies are static and do not respond to short-term variations in traffic noise emissions. Although variable speed limit (VSL) systems are widely deployed for safety and congestion mitigation, traffic noise is rarely treated as an explicit operational control objective.

This paper proposes a noise-aware VSL framework that integrates aggregated traffic-state estimation with a simplified CNOSSOS-EU–based emission indicator. A stage-based controller with time-varying reference thresholds dynamically adjusts discrete speed-limit levels in response to estimated emission conditions. The framework is evaluated using microscopic traffic simulation calibrated with empirical motorway data and replicated across multiple stochastic realisations.

Over a 24-hour evaluation period, the adaptive strategy reduces the receiver-based equivalent sound level by 2.9\,dB(A) relative to unrestricted traffic conditions, while maintaining an average vehicle speed approximately 11.3\,km/h higher than a permanently imposed low-speed regime. Period-wise analysis shows that speed reductions are activated selectively when emission levels approach calibrated targets, rather than enforcing a constant intermediate limit. Traffic stability indicators reveal moderate increases in speed variability compared with unrestricted operation, but substantially lower braking intensity than under uniform low-speed enforcement.

These results demonstrate the feasibility of integrating environmental performance indicators into operational speed control, providing a practical complement to conventional infrastructure-based noise mitigation measures.

\end{abstract}
\section{Introduction}
\label{sec:introduction}

Road traffic noise constitutes a persistent environmental burden in urban agglomerations and along major motorway corridors. The European Environment Agency reports that approximately 92 million people across Europe are exposed to day–evening–night noise levels exceeding 55\,dB(A)~\cite{EEA2025}, a threshold associated with adverse cardiovascular and psychological health outcomes~\cite{WHO2018,Hahad2025NoiseMentalHealth}. Beyond public health considerations, the mitigation of excessive road traffic noise presents regulatory and operational challenges for transport authorities  facing increasingly stringent environmental requirements.

Conventional mitigation measures, including noise barriers~\cite{Barros2024} and low-noise pavement surfaces~\cite{Piao2022}, are infrastructure-based and inherently static. While effective at specific locations, they are capital-intensive and difficult to adapt once deployed. These limitations motivate complementary operational strategies that can dynamically influence traffic-induced noise emissions.

Standardised emission models such as CNOSSOS-EU~\cite{CNOSSOS2012}, RLS-19~\cite{RLS19eng}, and the Federal Highway Administration Traffic Noise Model~\cite{FHWA_TNM} identify vehicle speed, traffic flow, and fleet composition as primary determinants of road traffic noise emission. Empirical evidence further confirms that reductions in operating speed lead to measurable decreases in traffic noise levels~\cite{Brink2022}. This established relationship highlights active speed management as a practical operational lever for mitigating traffic noise.

Variable Speed Limit (VSL) systems are widely deployed to enhance motorway safety and throughput~\cite{lee2006evaluation}, with strategies ranging from rule-based heuristics to predictive and learning-based approaches~\cite{coppola2023fuzzy}. When environmental objectives are incorporated, most VSL studies focus on fuel consumption~\cite{Khondaker2015} or air pollutant emissions~\cite{bel2013effects}. Despite its direct relationship with vehicle speed and its well-established emission modelling framework, traffic noise is rarely considered an explicit real-time operational objective.

Previous research has examined the acoustic implications of traffic management primarily through offline scenario analyses. These studies typically estimate noise from observed traffic states~\cite{Pascale2023RTNM,Fredianelli2022TrafficITS} or evaluate predefined control strategies in simulation~\cite{Ece2018Modeling}. The integration of standardised noise emission modelling into operational speed management with real-time speed updates remains limited.

This paper presents a noise-responsive VSL framework that integrates real-time traffic sensing with a simplified CNOSSOS-EU-based noise emission indicator to determine speed stages dynamically. The controlled motorway segment is partitioned into an upstream perception and adjustment zone and a downstream noise-sensitive target zone. The noise emission indicator is used at each control interval to select the appropriate speed stage based on observed traffic conditions.

The proposed framework is evaluated through microscopic traffic simulation calibrated using detector data from the German Federal Highway and Transport Research Institute (BASt) counting network. This calibration ensures realistic demand patterns and fleet compositions representative of real-world motorway traffic conditions. The simulation study enables a quantitative assessment of the trade-off between traffic performance and noise reduction under dynamic speed control.

The main contributions of this work are:
\begin{itemize}
\item Establishing an operational pathway for incorporating standardised traffic noise emission modelling into VSL management.
\item Providing quantitative evidence on the achievable noise reduction and associated traffic performance impacts under realistic motorway demand conditions.
\end{itemize}

\section{Methodology}
\label{sec:methodology}

\subsection{Framework Overview}
\label{sec:framework_overview}
\begin{figure}[htbp]
    \centering
    \includegraphics[width=1\linewidth]{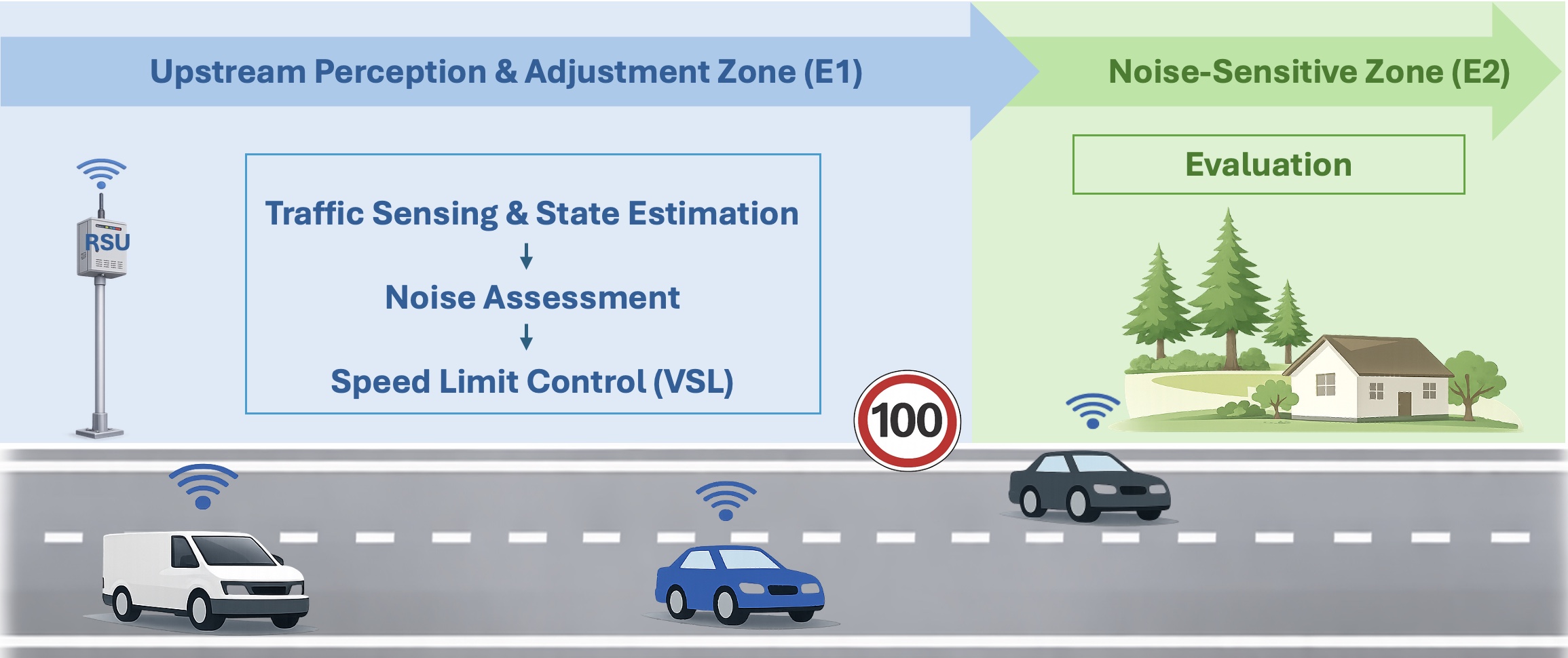}
    \caption{Overview of the noise-responsive VSL framework. The motorway stretch is partitioned into an upstream perception and adjustment zone ($E_1$) and a downstream noise-sensitive evaluation zone ($E_2$).}
    \label{fig:framework}
\end{figure}

The proposed framework consists of three functional components, as illustrated in Fig.~\ref{fig:framework}. The controlled motorway stretch is partitioned into an upstream perception and adjustment zone ($E_1$) and a downstream noise-sensitive evaluation zone ($E_2$), such that speed adaptations are largely completed before vehicles enter $E_2$. Within this structure, the \textbf{Traffic Sensing and State Estimation} component observes and aggregates vehicle-level traffic states at the entrance to $E_1$. These states are processed by the \textbf{Noise Emission Assessment} component to derive a segment-level emission indicator, which is subsequently used by the \textbf{VSL Control} component to periodically adjust the active speed limit.


\subsection{Traffic State Representation}
\label{sec:traffic_state}
Traffic-state information is assumed to be available via conventional roadside detection or connected-vehicle technologies. Vehicle observations at the entrance of $E_1$ are aggregated over a sliding window of length $T_{\mathrm{win}}$, updated every control interval $T_{\mathrm{ctrl}}$.

For each CNOSSOS vehicle category $m$, the time-averaged flow rate $Q_m$ (veh/h) is computed over $T_{\mathrm{win}}$. These aggregated flows, together with the fleet composition, form the traffic input to the emission model.

\subsection{Noise Emission Model}
\label{sec:noise_model}
CNOSSOS-EU is the harmonised European framework for environmental noise assessment. Its full source–propagation formulation is primarily intended for regulatory noise mapping. In the present study, a simplified emission indicator is derived from the CNOSSOS source model, retaining sensitivity to traffic flow, speed, and fleet composition, while omitting explicit propagation modelling.

\paragraph{Single-vehicle emission}

CNOSSOS-EU characterises vehicle noise by separating it into rolling and propulsion components. For vehicle category $m$ at speed $v$~(km/h), the total sound power level per octave band is

\begin{equation}
    L_{W,m}(v) = 10\log_{10}\!\left(
        10^{L_{W,R,m}(v)/10} + 10^{L_{W,P,m}(v)/10}
    \right),
\end{equation}
where $L_{W,R,m}(v)$ denotes the rolling noise component 
and $L_{W,P,m}(v)$ the propulsion noise component for 
vehicle category $m$, with coefficients defined in~\cite{CNOSSOS2012}. The A-weighted sound power level $L_{W,m,A}(v)$ is obtained through standard A-weighting and energetic summation across octave bands.

\paragraph{Segment-level aggregation}
For each vehicle category $m$, the directional line-source emission indicator is computed from the observed flow $Q_m$ and the corresponding mean operating speed $v_m$ as

\begin{equation}
    L_{W,\mathrm{eq,line},m}^{\prime}
    =
    L_{W,m,A}(v_m)
    +
    10\log_{10}\!\left(
        \frac{Q_m}{1000 \cdot v_m}
    \right).
\end{equation}

The total segment-level emission indicator is obtained by energetic summation over all vehicle categories:

\begin{equation}
    L_{W,\mathrm{eq,line}}^{\prime}
    =
    10\log_{10}\!\left(
        \sum_m
        10^{L_{W,\mathrm{eq,line},m}^{\prime}/10}
    \right).
\end{equation}

\subsection{Variable Speed Limit Control}
\label{sec:speed_regulation_strategy}
The VSL controller selects the active speed-limit stage such that the predicted noise emission indicator does not exceed the time-dependent reference $L_{\mathrm{target}}(t)$.

A finite ordered set of $N = 5$ discrete speed-limit stages is defined. Each stage $k$ prescribes the pair $\bigl(v_{\mathrm{pass}}(k), v_{\mathrm{hgv}}(k)\bigr)$, where the subscripts denote passenger vehicles and heavy goods vehicles (HGVs), respectively. Both limits are non-increasing in $k$. The stages are ordered such that the passenger-vehicle limits converge toward the HGV limits before both are reduced jointly (Table~\ref{tab:vsl_stages}). The CNOSSOS light-vehicle category and powered two-wheelers are assigned to the passenger-vehicle limit, whereas the medium-heavy and heavy categories are subject to the HGV limit.

Let $k_t$ denote the active stage at control interval $t$. At each interval, all candidate stages $k \in \{0, \ldots, N-1\}$ are evaluated by substituting the stage-implied speeds into the emission model defined in Section~\ref{sec:noise_model}, using the observed traffic flows $\{Q_m\}$. This yields the predicted emission level $L_{W,\mathrm{eq,line}}^{\prime(k)}$ for every evaluated stage.

The desired stage is then selected as the least restrictive stage that satisfies the emission constraint:
\begin{equation}
k_t^{\mathrm{des}} =
\min \left\{
k :
L_{W,\mathrm{eq,line}}^{\prime(k)}
\le L_{\mathrm{target}}(t)
\right\}.
\end{equation}

If no candidate stage satisfies the constraint, the most restrictive stage $N-1$ is applied.

To prevent oscillatory behaviour caused by short-term traffic fluctuations, the direction of the desired stage change must remain unchanged for \(N_{\mathrm{persist}}=5\) consecutive control intervals. The controller then moves one stage toward the desired stage.

During the core nighttime period outside the transition windows, the most restrictive stage is enforced directly. This reflects the stricter regulatory noise limits applicable at night~\cite{WHO2018} and avoids unnecessary control switching under very low traffic demand.


\section{Simulation and Evaluation}
\label{sec:simulation}
\subsection{Simulation Setup}
\label{sec:simulation_setup}
A 3\,km unidirectional three-lane motorway corridor is modelled in Simulation of Urban Mobility (SUMO) and partitioned into three equal 1\,km segments: warm-up, $E_1$ (perception/adjustment), and $E_2$ (evaluation). Geometry and demand are derived from BASt permanent counting data collected on the A66 motorway (R1) in 2023~\cite{BASt2023}. Hourly traffic volumes and BASt vehicle classes are mapped to the four standard CNOSSOS-EU categories (light, medium heavy, heavy, and powered two-wheelers), which are subsequently implemented in SUMO through hourly vehicle-flow definitions. While all four categories are retained in the emission calculations, speed limits are applied to two aggregated vehicle groups for control purposes.

\begin{figure}[htpb]
    \centering
    \includegraphics[width=0.8\linewidth]{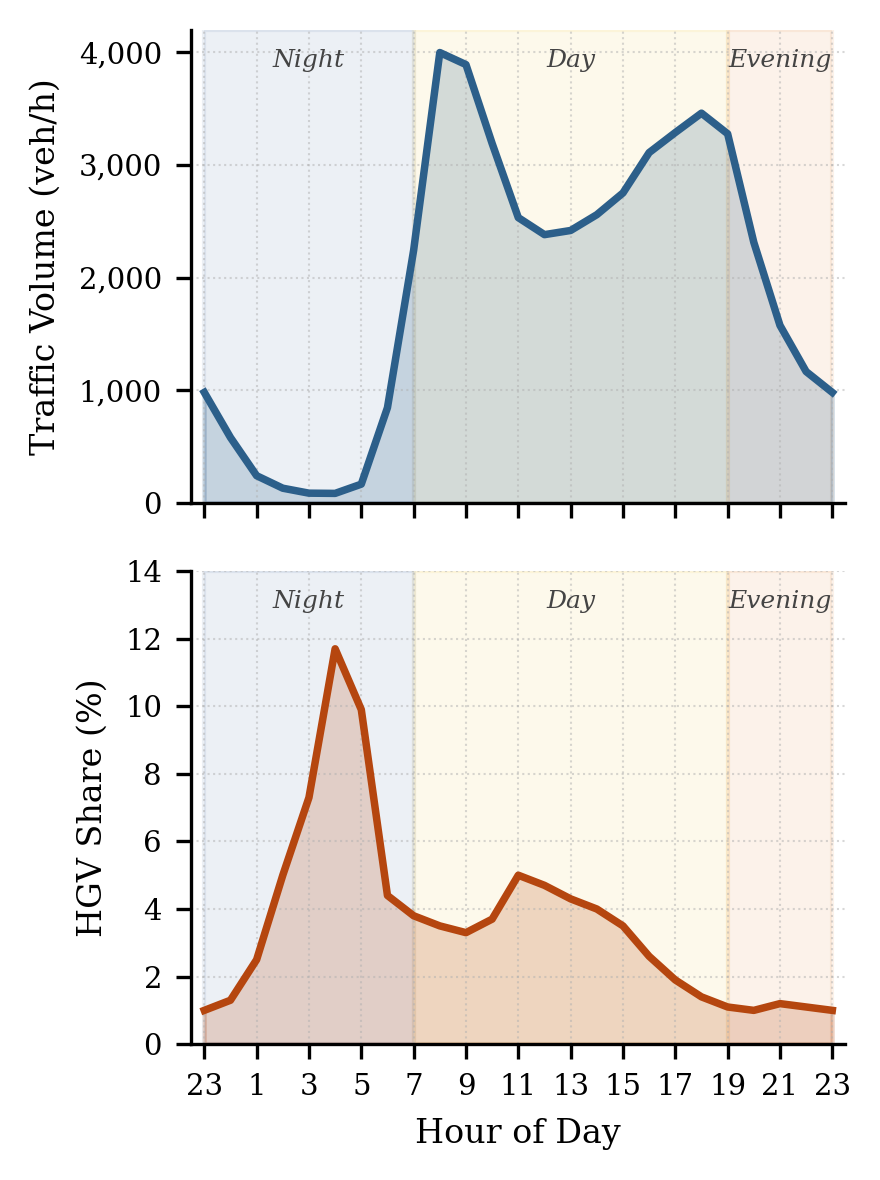}
    \caption{Diurnal traffic demand profile derived from BASt permanent counting data on the A66 (R1) in 2023. The primary y-axis shows total traffic volume (veh/h), while the secondary y-axis represents HGV share (\%).}
    \label{fig:demand_profile}
\end{figure}

Figure~\ref{fig:demand_profile} shows a typical diurnal traffic pattern with two daytime peaks and an increased HGV share during nighttime hours.

Driver heterogeneity is represented within the Krauss car-following model~\cite{Krauss1998} through three behavioural profiles: conservative (30\%), normal (50\%), and aggressive (20\%). All remaining model parameters follow the SUMO default settings.

The simulation spans 25 hours, starting at 06:00. The first hour is discarded to remove initialisation effects, yielding a 24-hour evaluation period. 
Lane-changing behaviour emerges endogenously from the implemented car-following and lane-changing models. 
Each scenario is replicated using five independent random seeds, and the reported results are averaged across all replications.

\subsection{Experimental Scenarios} \label{subsec:scenarios}
Three speed-limit control configurations are compared under identical network geometry, demand, and fleet composition. Together, they provide two reference endpoints and an adaptive policy to quantify the mobility--noise trade-off on the studied corridor.

\textit{S1 --- Baseline (Uncontrolled):} 
Speed limits remain at the maximum permissible values throughout the simulation (100\,km/h for passenger cars and 80\,km/h for HGVs). This scenario represents unconstrained operation and serves as the mobility reference.

\textit{S2 --- Dynamic VSL (Proposed):}
The noise-aware VSL controller described in Section~\ref{sec:speed_regulation_strategy} is activated. The controller dynamically selects the active speed-limit stage from Stage~0 (100/80\,km/h) to Stage~4 (60/60\,km/h), as summarised in Table~\ref{tab:vsl_stages}.

\textit{S3 --- Static-Low:} 
A uniform low-speed regime is enforced throughout the simulation (60\,km/h for both passenger cars and HGVs). This non-adaptive configuration represents a permanently noise-oriented policy and serves as a lower-noise reference under fixed-limit operation.

\subsection{Control Threshold Calibration}
\label{subsec:target_calibration}
The time-varying control threshold $L_{\mathrm{target}}(t)$ is derived from period-specific average emission levels computed from the observed traffic demand using the CNOSSOS-based emission model described in Section~\ref{sec:noise_model}.

For each assessment period $p \in \{\text{day}, \text{evening}, \text{night}\}$, the mean emission indicator $\bar{L}_p$ is calculated under baseline speed conditions (S1). The resulting baseline emission levels are:

\begin{itemize}
\item Day (07:00--19:00): $\bar{L}_{\text{day}} = 89.54$ dB(A)
\item Evening (19:00--23:00): $\bar{L}_{\text{evening}} = 88.10$ dB(A)
\item Night (23:00--07:00): $\bar{L}_{\text{night}} = 80.01$ dB(A)
\end{itemize}

The control target for period $p$ is defined as

\begin{equation}
L_{\mathrm{target},p}
=
\bar{L}_p - \Delta L.
\end{equation}

The value $\Delta L = 1.5$\,dB is selected to ensure that the controller operates within the effective range of the defined speed-stage set. A substantially larger value of $\Delta L$ would permanently activate the most restrictive stage, whereas a near-zero value would rarely trigger speed adjustments. The parameter remains configurable for corridor-specific requirements.

To ensure temporal continuity, the period-specific targets are connected by linear interpolation within symmetric transition windows of half-width $T_{\mathrm{smooth}}$, yielding a continuous piecewise-linear profile $L_{\mathrm{target}}(t)$.

\begin{figure}[htbp]
    \centering
    \includegraphics[width=\linewidth]{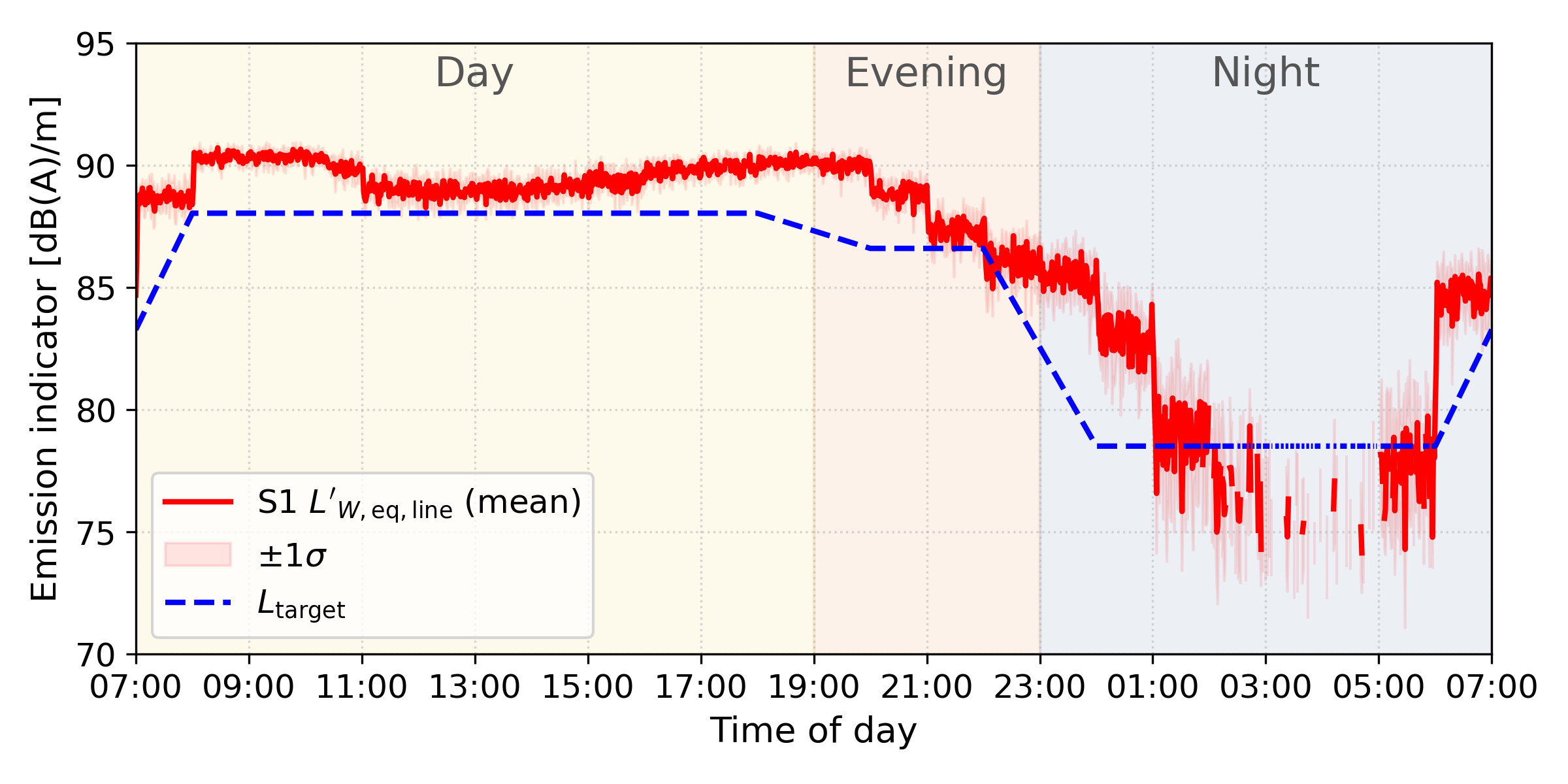}
    \caption{Simulated baseline emission indicator under Scenario~S1 (mean $\pm 1\sigma$ across five seeds) together with the calibrated time-varying control threshold $L_{\mathrm{target}}(t)$. Shaded areas indicate the assessment periods, and dashed segments represent smoothing transitions between periods.}
    \label{fig:target_profile}
\end{figure}

Figure~\ref{fig:target_profile} shows the temporal evolution of the baseline emission indicator under Scenario~S1 (mean $\pm 1\sigma$ across five seeds) together with the calibrated target profile.

\subsection{Control Parameter Settings}
\label{subsec:control_params}
Given the calibrated target profile, all controller parameters are fixed across scenarios and summarised in Table~\ref{tab:control_params}.

Traffic states are aggregated over $T_{\mathrm{win}}$, and the controller is executed every $T_{\mathrm{ctrl}}$. Stage updates follow the persistence condition defined in Section~\ref{sec:speed_regulation_strategy}. The parameter $T_{\mathrm{smooth}}$ corresponds to the transition window defined in Section~\ref{subsec:target_calibration}.

\begin{table}[htbp]
\centering
\caption{Controller parameter settings.}
\label{tab:control_params}
\begin{tabular}{lll}
\hline
Parameter & Symbol & Value \\
\hline
Observation window      & $T_{\mathrm{win}}$      & 60\,s \\
Control interval        & $T_{\mathrm{ctrl}}$     & 60\,s \\
Smoothing window        & $T_{\mathrm{smooth}}$   & 3600\,s \\
Reduction target        & $\Delta L$              & 1.5\,dB \\
Persistence requirement & $N_{\mathrm{persist}}$  & 5 intervals \\
\hline
\end{tabular}
\end{table}

\begin{table}[htbp]
\centering
\caption{VSL stage definitions.}
\label{tab:vsl_stages}
\begin{tabular}{ccc}
\hline
Stage & Passenger vehicles (km/h) & HGVs (km/h) \\
\hline
0 (baseline) & 100 & 80 \\
1            &  90 & 80 \\
2            &  80 & 80 \\
3            &  70 & 70 \\
4 (minimum)  &  60 & 60 \\
\hline
\end{tabular}
\end{table}

\subsection{Noise Evaluation Method}
\label{subsec:noise_evaluation}

Acoustic performance is evaluated using a receiver-based A-weighted sound pressure level metric within segment $E_2$. In contrast to the aggregated line-source emission indicator used for control, the evaluation is computed at microscopic resolution based on the instantaneous positions of individual vehicles.

A virtual receiver is located at the midpoint of segment $E_2$, 25\,m laterally from the roadway centreline and 4\,m above ground level. At 10\,s intervals, all vehicles within \(E_2\) contribute according to their instantaneous position, speed, and vehicle category.

For vehicle $i$ belonging to category $m$, the instantaneous A-weighted sound pressure level at the receiver is

\begin{equation}
L_{p,i} = L_{W,m,A}(v_i) - 20\log_{10}(d_i) - 11,
\end{equation}
where $d_i$ denotes the three-dimensional source-receiver 
distance, and the constant $-11$\,dB accounts for 
free-field spherical divergence.

Vehicle contributions are combined by energetic summation, and noise performance is quantified using the equivalent continuous sound level over a sliding 60\,s time window, $L_{\mathrm{eq},60}$.

Atmospheric absorption, ground effects, and shielding are neglected, and identical propagation assumptions are applied across all scenarios.

\section{Results}
\label{sec:results}

\subsection{Aggregate Performance Comparison}

Table~\ref{tab:overall} summarises the overall acoustic and traffic performance over the full 24-hour evaluation period.

\begin{table}[htbp]
\centering
\caption{Overall performance metrics averaged over five random seeds
(full 24-hour evaluation period).}
\label{tab:overall}
\begin{tabular}{lrrr}
\hline
Metric & S1 & S2 & S3 \\
\hline
$L_{\mathrm{eq}}$ (dB(A))       & 67.3 & 64.4 & 62.6 \\
$\Delta L_{\mathrm{eq}}$ vs S1  & --   & $-$2.9 & $-$4.7 \\
Avg.\ speed -- car (km/h)       & 94.0  & 68.6 & 57.2 \\
Avg.\ speed -- HGV (km/h)       & 78.9 & 67.2 & 56.0 \\
Avg.\ speed -- all (km/h)       & 93.5 & 68.5 & 57.2 \\
\hline
\end{tabular}
\end{table}

The uncontrolled baseline (S1) yields a 24-hour equivalent sound level of 67.3\,dB(A)
with a space-mean speed of 93.5\,km/h. The static low-speed configuration (S3)
achieves the largest acoustic reduction ($-4.7$\,dB) but reduces mean speed
to 57.2\,km/h.

The adaptive strategy (S2) lowers $L_{\mathrm{eq}}$ by 2.9\,dB(A) while maintaining an average speed of 68.5\,km/h. Thus, S2 provides substantial noise reduction while avoiding the full mobility penalty associated with a permanent low-speed regime.

\subsection{Period-wise and Temporal Analysis}
Table~\ref{tab:period} summarises the period-averaged receiver sound levels and space-mean speeds.

\begin{table}[htbp]
\centering
\caption{Period-wise receiver $L_{\mathrm{eq}}$ (dB(A)) and space-mean speed (km/h), averaged over five random seeds.}
\label{tab:period}
\begin{tabular}{llrrrr}
\hline
Period & Metric & S1 & S2 & S3 \\
\hline
\multirow{2}{*}{Day}
  & $L_{\mathrm{eq}}$ (dB(A)) & 69.0 & 66.2 & 64.3 \\
  & Avg.\ speed (km/h) & 92.4 & 69.0 & 56.9 \\
\hline
\multirow{2}{*}{Evening}
  & $L_{\mathrm{eq}}$ (dB(A)) & 67.7 & 64.9 & 62.9 \\
  & Avg.\ speed (km/h) & 96.0 & 69.6 & 57.8 \\
\hline
\multirow{2}{*}{Night}
  & $L_{\mathrm{eq}}$ (dB(A)) & 60.7 & 56.1 & 55.8 \\
  & Avg.\ speed (km/h) & 99.7 & 61.1 & 59.5 \\
\hline
\end{tabular}
\end{table}

During the daytime and evening periods, S2 reduces $L_{\mathrm{eq}}$ by 2.7\,dB and 2.8\,dB, respectively, relative to S1, while maintaining substantially higher
mean speeds than S3.

During nighttime conditions, the controller enforces the most restrictive stage. However, due to the smoothing and persistence mechanisms, S2 experiences gradual transitions at period boundaries. As a result, its period-averaged noise level
remains slightly higher than that of S3, which applies the low-speed limit continuously.

\begin{figure}[htbp]
    \centering
    \includegraphics[width=0.9\linewidth]{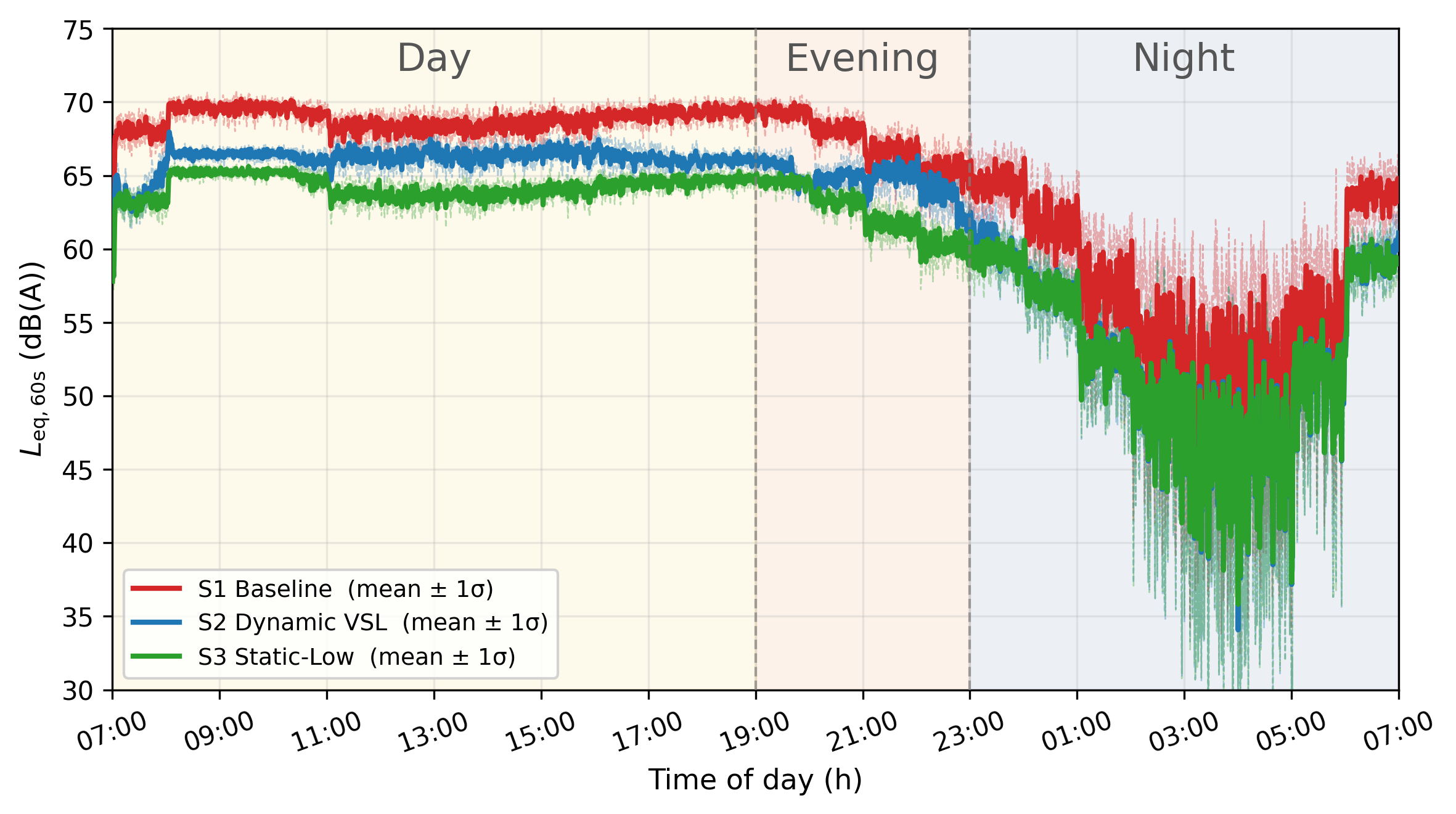}
    \caption{Time-series comparison of receiver sound pressure levels under dynamic and static speed control schemes during the evaluation period (mean $\pm1\sigma$ across five random seeds).}
    \label{fig:spl_timeseries}
\end{figure}

Figure~\ref{fig:spl_timeseries} illustrates the temporal evolution of the receiver sound pressure level (mean $\pm1\sigma$ across five seeds). The ordering $S1 > S2 > S3$ is preserved throughout the day,
confirming that the adaptive strategy consistently operates between the uncontrolled and static-low regimes.

Larger short-term fluctuations are observed during nighttime across all scenarios. These arise from low traffic demand, where individual vehicle passages dominate the 60\,s equivalent level, and therefore reflect statistical effects rather than control instability.

\subsection{Control Behaviour and Traffic Stability}
\label{subsec:control}

\begin{figure}[htbp]
    \centering
    \includegraphics[width=\linewidth]{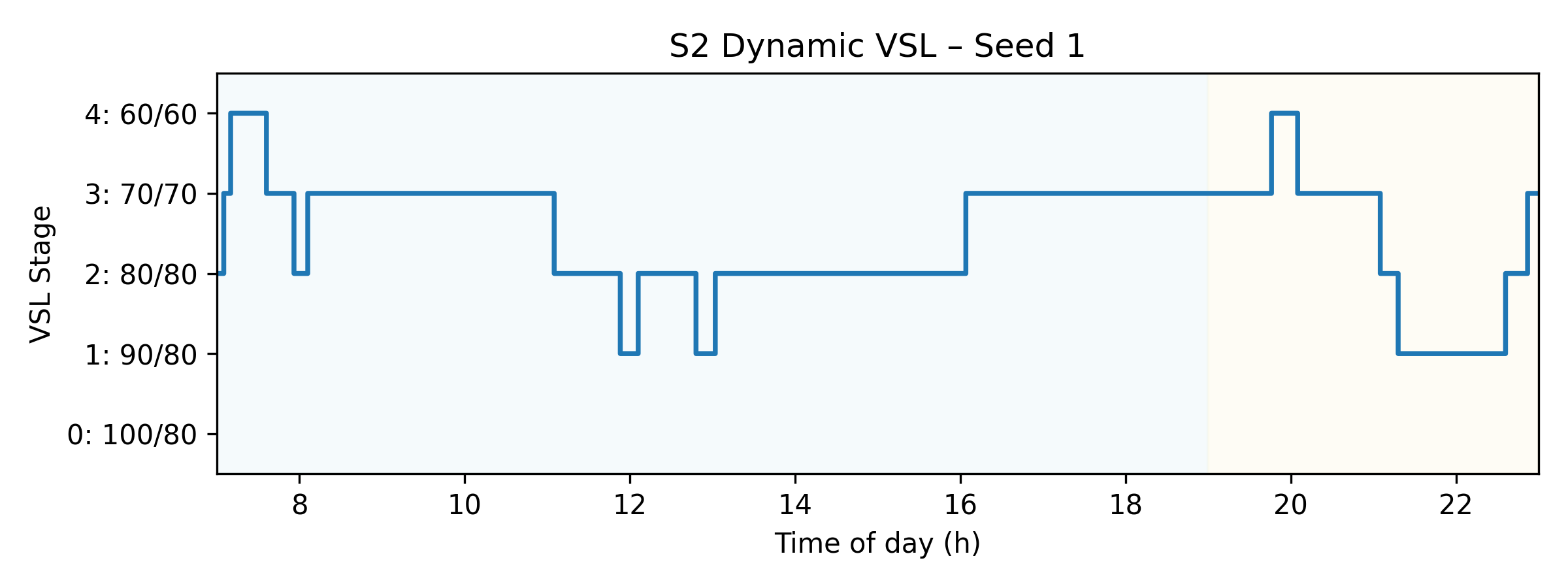}
    \caption{Example VSL stage evolution under S2 (representative seed), showing discrete speed-limit adjustments across assessment periods.}
    \label{fig:vsl_profile}
\end{figure}

Figure~\ref{fig:vsl_profile} illustrates the discrete stage evolution
of the adaptive VSL strategy (S2) for a representative simulation seed.
Stage changes occur in structured increments within the predefined stage set, without rapid back-and-forth switching, indicating stable discrete control behaviour. 

\begin{table}[htbp]
\centering
\caption{Stage distribution under S2 for daytime and evening periods}
\label{tab:vsl_distribution}
\begin{tabular}{lrrrrr}
\hline
Period & Stage 0 & Stage 1 & Stage 2 & Stage 3 & Stage 4 \\
\hline
Day     & 0.0\% & 1.0\% & 46.0\% & 48.4\% & 4.6\% \\
Evening & 0.0\% & 32.8\% & 14.0\% & 41.8\% & 11.4\% \\
\hline
\end{tabular}
\end{table}

While Figure~\ref{fig:vsl_profile} illustrates one representative control trajectory, Table~\ref{tab:vsl_distribution} summarises the operating distribution averaged across five runs.

During daytime, operation is mainly concentrated in Stages 2 and 3. In the evening, the stage distribution becomes more dispersed, with Stages 1 and 3 occurring most frequently. The mean number of stage transitions remains moderate (8.2 during daytime and 6.2 during evening), indicating stable control behaviour.

\begin{figure}[htbp]
\centering
\includegraphics[width=\linewidth]{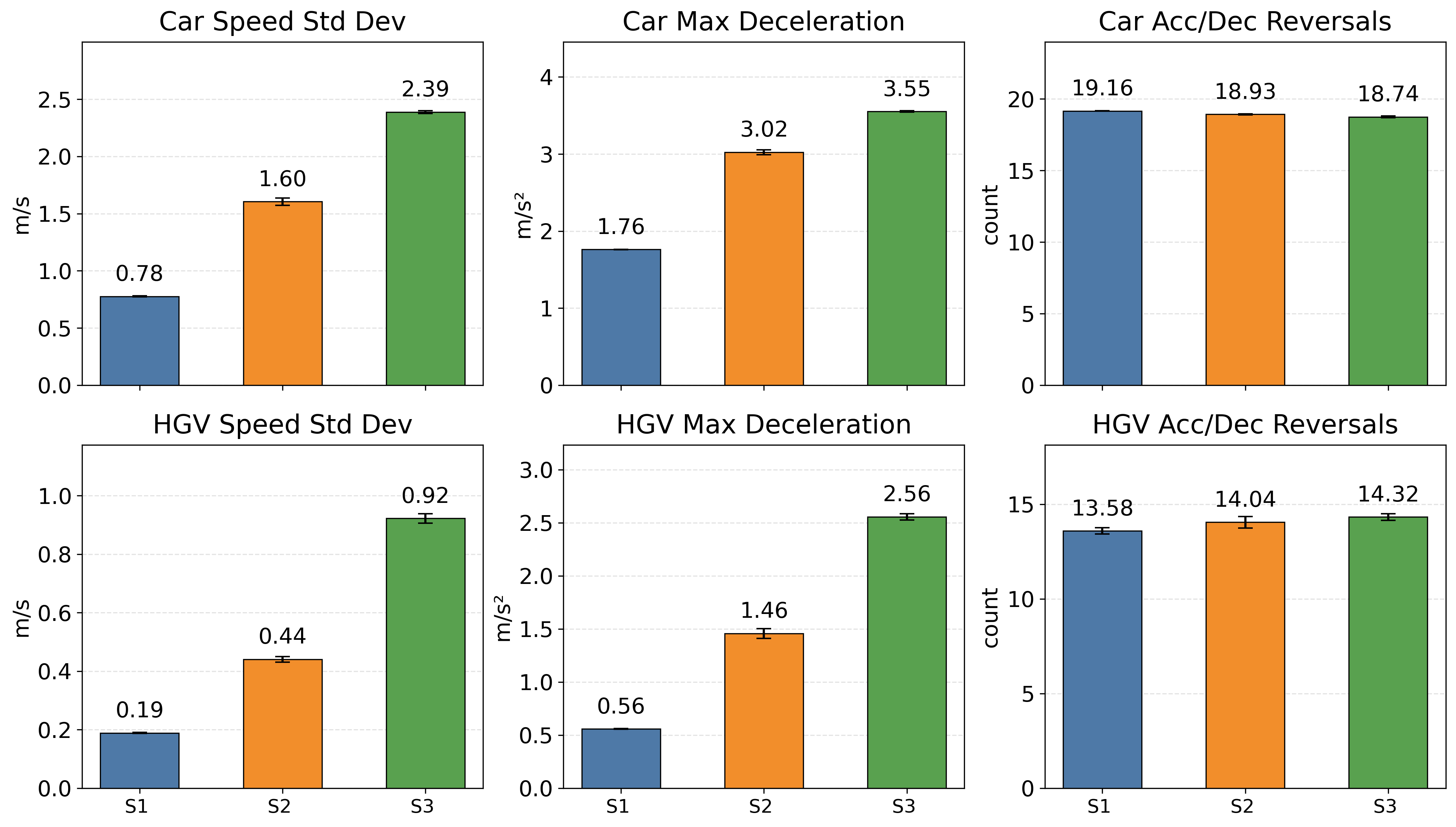}
\caption{Comparison of traffic stability indicators at segment $E_1$, averaged over five independent random seeds. Error bars represent the standard deviation across seeds.}
\label{fig:stopgo}
\end{figure}

To assess potential side effects of dynamic regulation, traffic stability indicators are evaluated at segment $E_1$ (Figure~\ref{fig:stopgo}). 
For passenger cars, speed standard deviation increases from 0.78\,m/s under S1 to 1.6\,m/s under S2, and 2.39\,m/s under S3. Maximum deceleration follows the same pattern (1.76, 3.02, and 3.55\,m/s$^2$, respectively). HGVs exhibit a similar monotonic trend, with speed standard deviation of 0.19, 0.44, and  0.92\,m/s and maximum deceleration of 0.56, 1.46, and 2.56\,m/s$^2$ under S1, S2, and S3, respectively.


These results indicate that although S2 introduces additional variability relative to unrestricted operation, the dynamic stage adjustments do not lead to excessive braking or pronounced stop-and-go behaviour. In contrast, the permanently imposed low-speed regime (S3) induces substantially stronger braking responses and larger speed dispersion.

Notably, the number of acceleration--deceleration reversals remains nearly unchanged across scenarios for both vehicle classes, suggesting that oscillatory behaviour arises primarily from the underlying car-following dynamics rather than from the VSL logic itself.

Overall, the adaptive strategy achieves noise mitigation with a moderate and controlled impact on traffic stability and remains clearly less disruptive than the static low-speed configuration.

\section{Conclusion and Discussion}
\label{sec:conclusion}
This study presents a noise-aware VSL framework that integrates a simplified CNOSSOS-EU-based emission indicator into operational motorway speed control. By embedding environmental performance metrics within the regulation logic, the framework enables real-time noise mitigation through adaptive speed management.

Simulation results demonstrate that dynamic speed regulation achieves a balanced trade-off between acoustic performance and traffic efficiency. Relative to the uncontrolled baseline (S1), the adaptive strategy (S2) reduces the 24-hour equivalent sound level by 2.9\,dB(A), while preserving an average speed approximately 11.3\,km/h higher than under the static low-speed configuration (S3). At the same time, S2 exhibits milder traffic stability impacts than permanent restriction.

These findings indicate that selective and time-varying speed reductions can mitigate traffic-induced noise without imposing a continuous mobility penalty. Compared with a uniform low-speed regime, adaptive regulation achieves substantial environmental benefits with reduced disruption to traffic dynamics.

Despite these promising results, several limitations remain.

\textit{Sensing and emission modelling.}
The current implementation relies on an aggregated CNOSSOS-based source formulation and assumes reliable traffic-state estimation. In practice, traffic noise emissions are influenced by additional factors not explicitly captured in the simplified model, including tyre characteristics, road surface conditions, powertrain configurations, and vehicle-specific operating behaviour. These factors may introduce deviations between predicted and realised emission levels. Future work should investigate the integration of vehicle-level information through vehicle-to-infrastructure and vehicle-to-vehicle communication to refine emission estimation and enable adaptive calibration of the noise model under varying environmental and fleet conditions.

\textit{Comparative evaluation and control formulation.}
The present study evaluates the adaptive strategy against an uncontrolled baseline and a static low-speed regime. Additional comparisons with traffic-oriented VSL controllers would provide further insight into how noise-aware control interacts with conventional traffic management objectives. Moreover, the current controller does not explicitly optimise overall system performance. Future work could extend the framework toward multi-objective formulations incorporating travel time, energy consumption, and air pollutant emissions --- integrating noise as one component among several operational objectives --- potentially within model predictive or learning-based control frameworks.

\textit{Scenario realism and network deployment.}
The evaluation is conducted on a single-corridor motorway case study with controlled boundary conditions. Validation under heterogeneous geometries, empirical demand variability, and coordinated multi-segment control is required to assess scalability and operational robustness in larger networks. In such settings, the exhaustive evaluation of all candidate stages may need to be replaced by more efficient search procedures.

Overall, the results demonstrate the feasibility of embedding environmental performance indicators within motorway traffic control and establish a practical basis for environmentally informed VSL strategies.


\section*{ACKNOWLEDGMENTS}
This publication is part of the "Tyre Road Noise" research project, which is funded by the German Federal Ministry for Digital and Modernization based on a decision of the German Bundestag (funding reference: 01F2269A). 

\bibliographystyle{IEEEtran}
\bibliography{root} 

\end{document}